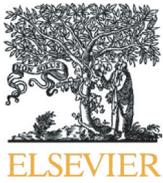
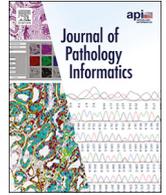

Review Article

# Seeing the random forest through the decision trees. Supporting learning health systems from histopathology with machine learning models: Challenges and opportunities

Ricardo Gonzalez [a,b,*,1], Ashirbani Saha [c,d], Clinton J.V. Campbell [e,f], Peyman Nejat [g], Cynthia Lokker [h], Andrew P. Norgan [i]

[a] *DeGroote School of Business, McMaster University, Hamilton, Ontario, Canada*
[b] *Division of Computational Pathology and Artificial Intelligence, Department of Laboratory Medicine and Pathology, Mayo Clinic, Rochester, MN, United States*
[c] *Department of Oncology, Faculty of Health Sciences, McMaster University, Hamilton, Ontario, Canada*
[d] *Escarpment Cancer Research Institute, McMaster University and Hamilton Health Sciences, Hamilton, Ontario, Canada*
[e] *William Osler Health System, Brampton, Ontario, Canada*
[f] *Department of Pathology and Molecular Medicine, Faculty of Health Sciences, McMaster University, Hamilton, Ontario, Canada*
[g] *Department of Artificial Intelligence and Informatics, Mayo Clinic, Rochester, MN, United States*
[h] *Health Information Research Unit, Department of Health Research Methods, Evidence and Impact, McMaster University, Hamilton, Ontario, Canada*
[i] *Department of Laboratory Medicine and Pathology, Mayo Clinic, Rochester, MN, United States*

| ARTICLE INFO | ABSTRACT |
|---|---|
| *Keywords:*<br>Pathology<br>Artificial intelligence<br>Machine learning<br>Learning health system<br>Image processing<br>Computer-assisted | This paper discusses some overlooked challenges faced when working with machine learning models for histopathology and presents a novel opportunity to support "Learning Health Systems" with them. Initially, the authors elaborate on these challenges after separating them according to their mitigation strategies: those that need innovative approaches, time, or future technological capabilities and those that require a conceptual reappraisal from a critical perspective. Then, a novel opportunity to support "Learning Health Systems" by integrating hidden information extracted by ML models from digitalized histopathology slides with other healthcare big data is presented. |

## Contents

* Corresponding author at: DeGroote School of Business, McMaster University, 1280 Main Street West, Hamilton, Ontario L8S4M4, Canada.
   *E-mail addresses:* gonzar5@mcmaster.ca, rgonzalez@lunenfeld.ca, (R. Gonzalez).
[1] Present address: Lunenfeld-Tanenbaum Research Institute, Mount Sinai Hospital, Joseph & Wolf Lebovic Health Complex, 881– 600 University Avenue, Toronto, Ontario M5G 1X5, Canada.

**Epigraph**

"We should take care not to make the intellect our goal. It has of course powerful muscles but no personality. It cannot lead, it can only serve, and it is not fastidious in its choice of a leader"—Albert Einstein, 1943.[1]

**Introduction**

Some months ago, when discussing the "gold-standard" in histopathology, the corresponding author [GR] shared an anecdote. As a pathology resident, he was expected to attend the National Institute of Legal Medicine and Forensic Science in Bogota (Colombia's capital and largest city) to improve his technical and analytical skills in forensic pathology. Many corpses of people who died of unnatural manners (i.e., homicides, suicides, accidents), suspiciously, unexpectedly, or for potential medical malpractices[2] arrived there daily. As in other countries, autopsies are part of some legal investigations there, and forensic pathologists' opinions are used to define responsibilities in courts[2,3] and are commonly awaited by the deceased's relatives and other concerned people.[4]

Every morning, 1 or 2 corpses were assigned to him. After reading a report with the circumstances of the death, which in many homicides and suicides were unknown (the sentence "no one saw or heard a thing" was commonly found), dissecting the cadaver to collect as much evidence as possible and discussing the relevant findings with his professor, the latter always ended up the conversation by asking the same question: "Okay, [GR] so… how are you gonna kill it?"

He was definitely not being trained to become a serial killer, and, as may be expected, some witnesses could testify on his behalf. Nevertheless, this question still resonates in his mind. And, it is not because he feels guilty about those people's deaths. He neither regrets stating the causes of death with the sometimes-limited evidence available (which was the real question behind his professor's daily pun). It is because that question reflects the uncertainty that pathologists may face in their routine practices since, for many diseases, their opinions are considered the "gold-standard" by social, or more specifically, medical conventions (i.e., those who capture the "ground truth").[5–10]

In addition to elaborating on the rationale behind accepting pathologists' opinions as the gold-standard, in the following paragraphs, the authors address other challenges faced by the machine learning (ML) community and pathologists working in artificial intelligence in histopathology. Also, a novel opportunity to support "Learning Health Systems"[11] with digital pathology by integrating hidden information extracted with ML models from histopathology slides with other healthcare big data is discussed at the end.

**Challenges**

Besides the need to increase adherence to reporting guidelines[12–18] to facilitate the understanding and comparison of experimental results[19–23] and other issues related to models' explainability[24] and data, model development, and model deployment,[25] some important challenges that have probably been overlooked are discussed below. They are separated into 2 groups based on their potential mitigation strategy.

Challenges that may need innovative approaches, time, or future technological capabilities:

*Expanding recognition capabilities of ML models*

Although the ability to recognize all the expected and unexpected diseases that could be diagnosed in any given histopathological case may be out of the scope of all or almost all ML models, this limitation certainly limits their applicability in routine clinical practices.

As explained below, ML models' outputs are limited by the examples used to train them.[26–29] Therefore, a reasonable approach to address this challenge is to create large and diverse but also more granular and comprehensive training datasets.[30–32] This process could be increasingly facilitated with a broader adoption of digital pathology and artificial intelligence platforms in the future[33–35] and collaborative efforts, such as those based on federated learning.[36,37]

*Extracting all the relevant information from Whole-Slide Images (WSIs)*

Largely due to computational constraints to processing complete WSIs, ML models have usually been trained using sampling methods that extract small patches obtained with one or a limited number of scanning magnifications.[38–44] The limitations of using these sampling strategies could be illustrated in the parable of the "blind men and an elephant".[45] As ML models' pattern recognition/prediction capabilities rely on the data contained in these patches, developing efficient methods to facilitate the extraction of all the relevant information from WSIs is of paramount importance.

Self-attention methods and other innovative approaches[44,46–52] as well as future improvements in computational power,[38–43,51,52] may help in this regard.

Challenges that may need a conceptual reappraisal from a critical perspective:

*Obtaining the "ground truth" for diagnostic/classification purposes in histopathology*

The reliability (i.e., "agreement," "reproducibility," or "inter- and intra-observer variability")[53] of pathologists' opinions has been a matter of concern for the ML community.[10,54] That is because pathologists' opinions are needed to train and test ML models for diagnosis/classification purposes, as they are regarded as the "gold-standard".[5,6]

As ML outputs might appear magical for people without domain knowledge of the field,[55] it could be understandable if pathologists' opinions look magical to non-pathologists and their cognitive processes are perceived as "black boxes".[56–58] However, contrary to what people may believe or expect, pathologists are not trained to become part of a selected group of legally accountable medical doctors to whom the "ground truth" is revealed to allocate diseases' names in patients' reports.[59] Contrarily, inter- and intraobserver variability can be regarded as inherent to these evaluative processes.[53,58,60,61] And, these processes may include some quantitative tasks, such as counting and measuring, in which computers can outperform pathologists when specific conditions are met.[8,62]

Although the best way to capture the "ground truth" to improve patients' health outcomes may remain an open discussion that could be framed in an ontological and epistemological manner[63,64] and acknowledging that resolving the historical debate between "naturalists" and "normativists"[65–68] is beyond the scope of this paper and the authors' philosophical argumentative capabilities, some of their insights are presented below. They are related to the way pathologists make diagnoses. With them, authors intend to help readers embrace the idea that, in practice and despite their limitations, more reliable sources of "ground truth" than pathologists' opinions may not be found for diagnostic/classification purposes in histopathology. Also, and hopefully, that everyone's beliefs or expectations need to be adjusted accordingly.

During their post-graduate training, pathology trainees are exposed to a large number of histopathology cases under the supervision of their future colleagues.[69,70] Besides refining their visual recognition capabilities,[69,70] they are expected to learn a conceptual framework and the field-specific terminology to make diagnoses.[59,60,70–75] Once graduated, with the proper clinical/surgical data and relevant auxiliary tests' results (e.g., obtained with immunohistochemical or molecular studies), pathologists make diagnoses by comparing the visual information they extract from patients' tissues/samples against sets of diagnostic/classification criteria (i.e., only after confirming that a tissue/sample meets some diagnostic/classification criteria do pathologists assign a disease name/classification category to it).[56,60,76,77] These sets of diagnostic/classification criteria are listed in histopathology classifications such as the WHO Classification of Tumours[78] and generally use concepts[60,79,80] to describe the presence/absence and the spatial distribution of some normal/abnormal cells and tissue components.[56,81,82] As histopathology classifications are typically created by pathologists to be used by pathologists,[78] only cells/tissue components expected to be recognizable by them are included in the diagnostic/classification criteria.[83,84] As a consequence, as long as the morphological classifications made (and to be used) by pathologists are needed to guide clinical decisions, pathologists' opinions will be regarded as the most reliable source of "ground truth" for diagnostic/classification purposes (i.e., the "gold-standard").[5,85]

Although a new potential role of "hidden" information extracted from WSIs in future healthcare systems will be addressed below, employing diagnoses made by consensus[6,10,86–88] and/or from pathologists with relevant expertise[10,89] seems to be a reasonable approach to train ML models as it reflects the best available standard of care.[10]

*Differentiating "hidden" from "diagnostically relevant" information*

WSIs carry large amounts of valuable information.[90] Some of this information has been visually recognizable for years and used by pathologists for diagnostic/classification purposes and, secondarily or indirectly, to predict prognosis and treatment outcomes.[54,77,90] Other information, previously hidden from human eyes, can now be extracted using some ML models.[41,91,92] This has created many opportunities with the potential to augment human skills.[93] For example, hidden information has demonstrated to be very powerful in predicting prognosis, treatment responses, and biomarkers using clinical follow-up data as the gold-standard[41,91,92,94–99] and in indexing and searching for cases with similar morphological patterns within large archives of WSIs.[100] However, an essential distinction between hidden and diagnostically relevant information must be made. As explained above, pathologists make diagnoses by comparing the visual information they extract from patients' tissues/samples against sets of diagnostic criteria listed in histopathology classifications.[56,76,77] As these sets of diagnostic criteria are created by pathologists to be used by pathologists,[78] only those cells/tissue components expected to be recognizable by pathologists are included in them.[83,84] Therefore, information hidden from pathologists' eyes that is not useful to assess if diagnostic/classification criteria are met (even if valuable for other purposes) can be considered irrelevant for morphologic diagnostic/classification purposes.

Nevertheless, developing ML models for morphologic diagnostic/classification purposes that recognize diagnostically relevant cells/tissue components may have some practical advantages. For example, it could be a reasonable approach to make ML models more applicable in the long term. That is because histopathology classifications usually change over time (i.e., are "moving targets"), but most diagnostically relevant cells/tissue components do not.[101–111] Also, as some of these cells/tissue components are listed in the different diagnostic criteria of many related diseases, these models could accelerate the development of tools that can support pathologists in more than one narrow task (e.g., by facilitating the recognition of several diseases).[112] In addition, the need to develop "explainable" algorithms would be less relevant and could gain regulatory agencies' approval easier if their scope is to be used as decision-support tools to improve pathologists' workflows (and not replace them in clinical settings by making diagnoses directly).[112–115] Lastly, better performances could be achievable with less effort with some commonly used ML models (i.e., convolutional neural networks). That is because cells/tissue components have some visual attributes (AKA features), such as colors, shapes, and textures, that these models can easily identify using convolutional kernels (i.e., feature extraction filters).[116,117] In contrast, to recognize all the specific diseases/classification categories while ruling out others with morphological similarities, ML models would need to assess if all the diagnostic/classification criteria are met. And, as there is still a large gap between ML models' pattern recognition and human-level concept learning,[118] the ability to understand all the concepts included in diagnostic/classification criteria may be beyond current ML models' capabilities. Noteworthily, concept-based explainable methods[119,120] could facilitate a rigorous assessment of this capability by pathologists' in the future.

*Going beyond external validation in clinical settings: The need for iterative model retraining[26,121–123] after ML models' deployment*

As specific patterns learned from biased-training datasets are not expected to improve the performance of ML models when tested with independent datasets, only external validation is considered important evidence of generalizability.[124] However, obtaining good results in one or more external validation datasets doesn't guarantee that the model will perform well in all other settings.[26,32,125] As ML models' prediction/recognition capabilities are restricted by the amount and diversity of examples used to train them, patterns not properly represented in training datasets are not expected to be adequately predicted/recognized in validation datasets.[26–29] The ability to use pathologists' conceptual frameworks[60,73,79,80] (in addition to pixels-derived recurrent patterns learned from training datasets) could be needed if widely generalizable ML models are expected to be developed.[118,126,127] And even if, for practical reasons, it is assumed that future events (such as those expected to be predicted by ML models) will always resemble past events (e.g., those used to create training datasets), it is important to be mindful that "universally" generalizable models may still be unachievable if the generalization problem is approached from a philosophical perspective and other problems of induction, as those explained by Lauc,[128] are contemplated.

As stated by the FDA, validation datasets must contain sufficient cases representative of those the product will likely encounter during its intended use.[129] Therefore, for real-life practice, comprehensive validation datasets would need to be created for each clinical setting where models are planned to be used.[32] In addition, to improve their site-specific performances, datasets to update them (by retraining and/or fine-tuning their hyperparameters when the selected metrics are below a predefined threshold) would also need to be constructed.[32] This can become an iterative process for each institution, considering that the ML model's performance would need to be monitored as new cases with previously unseen relevant characteristics would permanently arrive to be assessed.[26,121–123,130–132] Although the iterative nature of this process may not make ML models "universally" generalizable,[26,125,128] it would certainly boost their learning capabilities by leveraging their ability to falsify prediction rules that lack empirical adequacy (as postulated by Buchholz and Raidl).[133] If some major technical challenges are overcome,[134–137] and these steps can be done automatically,[122,132,138,139] a site-specific autonomous endless self-learning process could eventually be developed.

**Opportunities**

Lastly, a novel opportunity to support "Learning Health Systems" with digital pathology will be discussed.

Pathologists have been creating, refining, and validating histopathology classifications for decades. Like other taxonomists,[85] some prefer the use of broad categories (so-called "lumpers"), and others favor more granularity ("splitters").[140–142] While broader categories may sometimes contribute to reducing pathologists' intra- and interobserver variability,[143–148] more granularity could be useful to separate groups of individuals with relevant differences for research purposes (e.g., to assess specific treatment responses) or to guide clinical decisions (e.g., to provide personalized treatments when differences in treatment-responses have been found).[149–151] However, disregarding preferences in terms of lumping or splitting and clinical pertinence or reproducibility of current histopathology classifications, it is still not uncommon to find cases that do not fit in any of their diagnostic categories.[74,152–155] When facing the associated uncertainty, pathologists could opt to request second opinions,[86–88,156] perform ancillary tests,[7,9,157,158] select the diagnostic category where the most significant findings fit,[159–161] write comments (e.g., recommending complementary studies/procedures or suggesting to correlate their findings with other data), or use descriptive terms or unstandardized diagnoses.[152–155,159] These strategies can help to reduce the risk of harming patients. However, the need to use them may partially explain why most histopathology classifications are updated recurrently, even if some of the tissue morphological changes used in their diagnostic criteria remain unchanged.[101–108]

Similar to what was stated by Dr. Juan Rosai more than 2 decades ago, there are still no techniques more cost-effective, flexible, and rapidly informative than the morphological assessment of tissues by pathologists in clinical settings.[96] However, some techniques whose original role was to support or complement morphological histopathology classifications (e.g., immunohistochemistry and omics-based studies)[9,77,90,110,158] are now redefining some of them[103,107,109] and in a few specific cases, replacing them for treatment purposes,[162,163] somewhat reminiscent of how microscopic morphologic assessments once started to improve the recognition and prediction capabilities of gross descriptions and clinical findings.[158] Although it may be impossible to anticipate how histopathology classifications will continue to evolve, as explained below, digital pathology brings a novel opportunity to integrate large amounts of histopathology information with other healthcare big data to improve the future provision of health services.

Even if the best methods to capture the ground truth remain debatable, and the problems of induction explained by Lauc[128] are ignored, the intrinsic ability of ML to falsify prediction rules that lack empirical adequacy[133] strengthened by the increasing availability of big data[13,164–166,229] could be leveraged to develop ML models that continuously integrate and assign specific weights (i.e., relative importance) to personal (e.g., clinical, radiological, histopathological, laboratory medicine, multi-omics, self-reported, and collected with wearable devices) and population-based empirical data (e.g., related to "social determinants of health")[97,167–171,229] to predict health outcomes dynamically.[122,123,139,172–174] In some of these models, hidden information extracted with ML models from WSIs that have shown to be valuable for prediction purposes[41,91,92,94–98] will most likely obtain high weights. And also, diagnostically relevant morphological information extracted by pathologists and pathologists' diagnoses based on histopathology classifications, validated for decades to predict health outcomes and to develop countless treatments,[54,77,90,150] will certainly get high weights in many of them.

Although major challenges related to data governance/management,[135,175–183] ethical/legal,[184–187] and environmental[188,189] considerations would need to be addressed, these "dynamic multimodal ML models" may become cost-effective in different populations[169,190–193] if conceived as the integrative tools needed to provide precision health[22,164,165,168,194–196] and support some of the iterative cycles of knowledge generation and continuous improvement of "learning health systems".[11,197–200]

### Final remarks

Finally, considering that ML capabilities are limited by the data used to train them[26–29] and that a significant amount of medical data currently comes from non-marginalized populations, it is highly predictable that some of these multimodal ML models will not work well on marginalized groups.[166,185,201–205] As this situation may perpetuate and even reinforce health disparities,[166,185,187,202,206–210] in addition to employing some debiasing methods,[202,203,206,211] allocating more resources to collect data from marginalized populations[212–214] and training ML models to address the social determinants of health or mitigate their effects[171,215–219] need to be contemplated. Also, regarding the epigraph, what Albert Einstein once said[1] about the intellect could also be applied to artificial intelligence algorithms nowadays. Accordingly, beyond any technical considerations, an over-reliance on ML models in health systems should be avoided[204] and the active participation of healthcare professionals during their developing, deploying, and monitoring could be beneficial.[220,221]

Even though the real impact of these "dynamic multimodal ML models" would undoubtedly need to be permanently assessed for specific health outcomes in different settings,[222–224] if they can help to create more equitable healthcare systems continuously supported by high-quality big data, it could be expected that the number of corpses that arrive at forensic pathology facilities will one day be reduced and that fewer of them will be assigned with preventable causes of death. Also, and hopefully, that people will only remember with a smile the uncertainty experienced by pathology residents when facing [GR]'s forensic pathology professor's puns.

### Funding

This "viewpoint" article did not receive any specific grant from funding agencies in the public, commercial, or not-for-profit sectors.

### Acknowledgments

None.

### Appendix A. Glossary of terms

| Glossary of terms | |
| --- | --- |
| Gold-standard | It is "the practical standard that is used to capture the ground truth".[5] It "may not always be perfectly correct, but in general is viewed as the best approximation".[5] |
| Ground truth | "A category, quantity, or label assigned to a dataset that provides guidance to an algorithm during training."[5] It is an "abstract concept of the truth".[5] |
| Digital pathology | "A blanket term that encompasses tools and systems to digitize pathology slides and associated meta-data, their storage, review, analysis, and enabling infrastructure".[5] |
| Whole-Slide Images | Digital representation of pathology glass slides at a microscope resolution. They are produced using slide scanners[5] |
| Artificial intelligence | A branch of computer science concerned with understanding and building intelligent entities (i.e., machines able to adapt to new situations).[99,225] |
| Machine learning | A subfield of AI that attempts to generate models that learn to make predictions on new data based on experience.[225,226] |
| External validation | Model evaluation conducted with data extracted from independent datasets.[14,124,227,228] They can be extracted from a different setting or source; for example, another clinic or hospital system[14,26,124,227,228] or from the same location but at a different point in time.[26] |